\def\year{2017}\relax
\documentclass[letterpaper]{article}
\usepackage{aaai17}
\usepackage{times}
\usepackage{helvet}
\usepackage{courier}
\usepackage{graphicx}
\usepackage{amssymb,amsmath,bm}
\usepackage{tabularx,threeparttable}
\usepackage{booktabs}
\usepackage{array,multirow}
\begin{document}
%
\title{Fine-grained Recurrent Neural Networks for Automatic Prostate\\ Segmentation in Ultrasound Images}
\author{
	Xin Yang$^*$, 
	Lequan Yu$^*$, 
	Lingyun Wu$^{\S}$,
	Yi Wang$^*$, 
	Dong Ni$^{\S}$,
	{\bf Jing Qin$^{\dagger}$},
	{\bf Pheng-Ann Heng$^{*,{\ddagger}}$} \\
	$^*$Department of Computer Science and Engineering, The Chinese University of Hong Kong\\
	$^{\S}$National-Regional Key Technology Engineering Laboratory for Medical	Ultrasound, \\
	School of Biomedical Engineering, Shenzhen University, China\\
	$^{\dagger}$Centre for Smart Health, School of Nursing, The Hong Kong Polytechnic University\\ 
	$^{\ddagger}$Guangdong Provincial Key Laboratory of Computer Vision and Virtual Reality Technology,\\
	Shenzhen Institutes of Advanced Technology, Chinese Academy of Sciences, China\\ 
	\texttt{\{xinyang,lqyu,ywang,pheng\}@cse.cuhk.edu.hk}
}

\maketitle
\begin{abstract}
	Boundary incompleteness raises great challenges to automatic prostate segmentation in ultrasound images. Shape prior can provide strong guidance in estimating the missing boundary, but traditional shape models often suffer from hand-crafted descriptors and local information loss in the fitting procedure. In this paper, we attempt to address those issues with a novel framework. The proposed framework can seamlessly integrate feature extraction and shape prior exploring, and estimate the complete boundary with a sequential manner. Our framework is composed of three key modules. Firstly, we serialize the static 2D prostate ultrasound images into dynamic sequences and then predict prostate shapes by sequentially exploring shape priors. Intuitively, we propose to learn the shape prior with the biologically plausible Recurrent Neural Networks (\textit{RNNs}). This module is corroborated to be effective in dealing with the boundary incompleteness. Secondly, to alleviate the bias caused by different serialization manners, we propose a multi-view fusion strategy to merge shape predictions obtained from different perspectives. Thirdly, we further implant the RNN core into a multiscale Auto-Context scheme to successively refine the details of the shape prediction map. With extensive validation on challenging prostate ultrasound images, our framework bridges severe boundary incompleteness and achieves the best performance in prostate boundary delineation when compared with several advanced methods. Additionally, our approach is general and can be extended to other medical image segmentation tasks, where boundary incompleteness is one of the main challenges. \par
\end{abstract}

\section{Introduction}
Prostate cancer is the one of the most common noncutaneous cancer in men around the world. The routine clinical modality for imaging the prostate is medical ultrasound. Segmenting prostate from ultrasound images is of essential importance for prostatic disease diagnoses and therapeutic choices, such as creating patient-specific anatomical models for surgical planning \cite{wang2016towards} and image-guided biopsy, real-time guidance for the placement of biopsy needles towards lesions \cite{hodge1989random}, and volumetric measurement for prostate shape evaluation \cite{terris1991determination}. However, manual delineation of prostate boundary is tedious, time-consuming and often irreproducible, even for experienced physicians. \par

\begin{figure}
	\centering
	\includegraphics[width=1.0\linewidth]{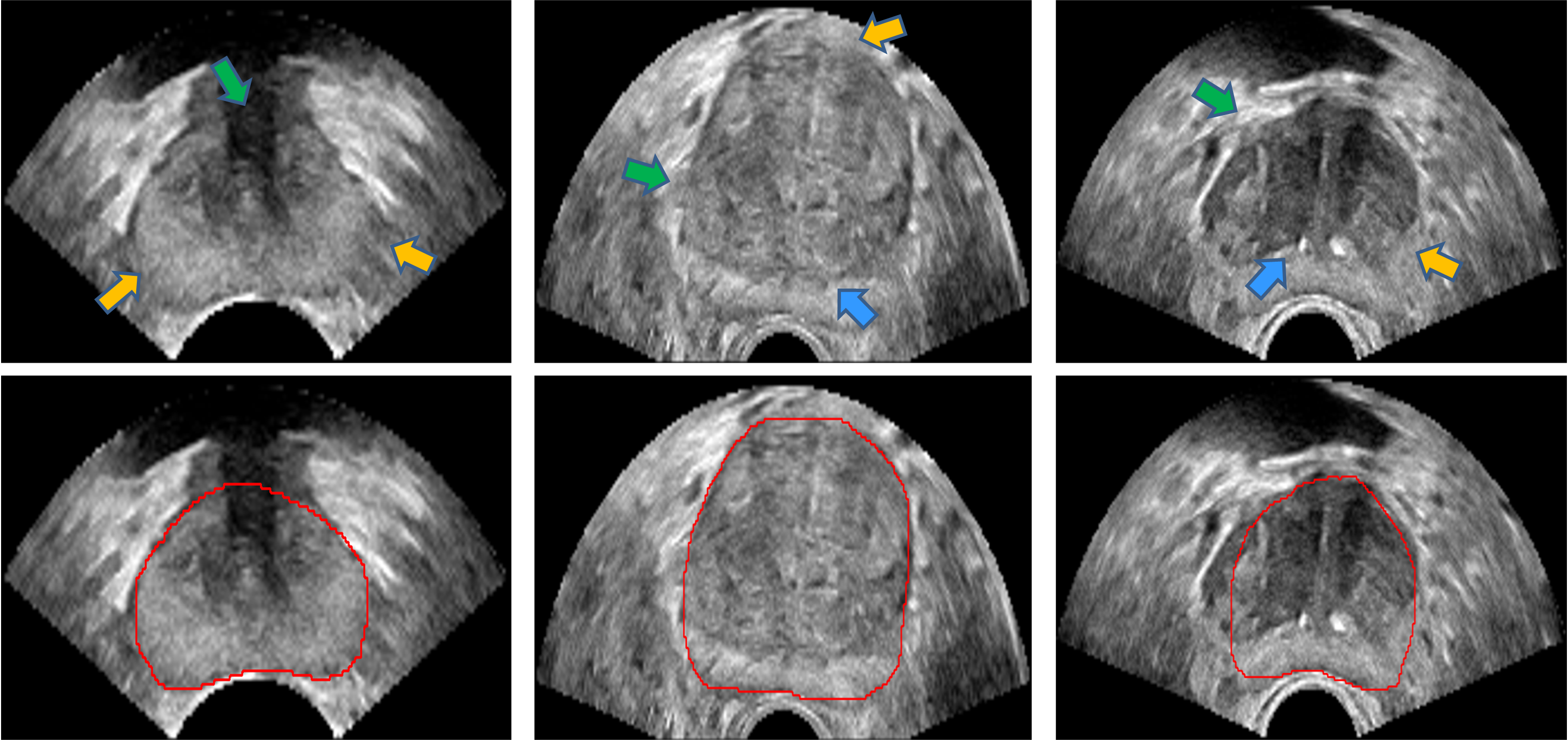}
	\caption{Boundary incompleteness in prostate ultrasound images. Green, yellow and blue arrows denote deficient boundary, ambiguous boundary and severe heterogeneity, respectively. Red curve denotes the segmentation ground truth.}
	\label{fig:incompleteness_show}
\end{figure}

Automatic solutions for accurate and efficient prostate segmentation in ultrasound image are highly desired. However, developing such automatic solutions remains very challenging for several reasons, as illustrated in Figure \ref{fig:incompleteness_show}. Firstly, different ultrasound images present diverse intensity distributions due to different imaging parameters, such as focus depth, Time Gain Compensation (TGC) and scanning orientation. Secondly, typical factors in ultrasound, including signal dropout, speckle noise, acoustic shadow and low contrast against surrounding tissues, cause the ambiguity, poor visibility and long-span occlusion in prostate boundaries \cite{noble2006ultrasound}. Thirdly, large variances in appearance, shape and size are often observed in prostates from different patients. Even the tissues belonging to a same prostate often present severe heterogeneity.

In the last decade, two main methodological categories have been studied as the solutions for automatic prostate segmentation:
\begin{itemize}
	\item \textit{Bottom-up fashion:} Methods in this stream mainly resort to classify each pixel as foreground (prostate tissue) or background with supervised classifiers. A set of support vector machines (SVM) associated with Gabor filters are utilized for prostate segmentation in ultrasound images \cite{zhan2006deformable}. Ensemble classifiers, such as Random Forests, have been explored to segment prostate in MRI and trans-rectal ultrasound \cite{mahapatra2015visual,ghose2013supervised}. However, it's still hard for those methods to estimate the missing boundary, since no constructive representations can be effectively collected in those ambiguous and long-span occluded regions. Based on considerably stable context information, regression based methods are established in recent years. By building a direct mapping between image appearance and its distance to boundary, Regression Forests are utilized for prostate segmentation in MRI images \cite{gao2014learning}. However, the assumption of stable context becomes less feasible in ultrasound images (Fig.~\ref{fig:incompleteness_show}). Recently, deep neural networks (DNNs), especially the Convolutional Neural Networks with hierarchical feature extraction capability are relighted and have achieved promising results in many vision tasks. DNNs are exploited to extract features which are more representative and effective than hand-crafted feature for prostate segmentation in MRI \cite{guo2016deformable}. The Fully Convolutional Network (FCN) \cite{long2015fully} characterized by an effective end-to-end prediction manner proves to be tractable for fetal ultrasound image segmentation \cite{chen2016iterative}. Discriminative as convolution based neural networks are, vanilla DNNs are bottlenecked in reasoning arbitrary-sized blind spots along object boundary \cite{li2016amodal}, and thus often output unexpected shape estimations.
	
	\item \textit{Top-down fashion:} This stream mainly explores the potential of global shape prior in guiding prostate segmentation. As a pioneer work, Active Shape Model (ASM) \cite{cootes1995active} illustrates its capacity in capturing both the shape and appearance variances for object segmentation. ASM equipped with Gabor descriptors is proposed for prostate segmentation in ultrasound images \cite{shen2003segmentation}. To enhance the ASM with more representative features, statistical analysis is performed on Gaussian derivatives of local histograms to learn the most informative descriptors for each landmark \cite{van2002active}. To tune the shape model with more precise displacements in fitting procedure, Robust Active Shape Model \cite{rogers2002robust} is proposed to discard displacement outliers in the image, which is subsequently developed for ultrasound image segmentation \cite{santiago20152d}. Discovering the low-rank property of similar shapes, an extra consistency constraint is developed to make shape model more robust to boundary deficiency in ultrasound images \cite{zhou2013active}. Partial active shape model has been used to estimate the missing boundaries of prostate in ultrasound shadow areas \cite{yan2010discrete}. Despite the difference between variations, the core components of most shape models are two: 1) capturing the main mode of shape variability by analyzing hand-crafted descriptors of a series of landmarks, which proves to be less tractable for prostate ultrasound image; 2) fitting the shape model to unseen occasions by minimizing specific cost functions, which is likely to be disturbed by local information loss when faced with arbitrary sized boundary deficiency \cite{zhou2013active}. \par
\end{itemize}

The \textit{bottom-up} fashion provides detailed prediction for each pixel in image, whereas suffers the lack of global shape prior in tackling boundary information loss. In contrast, shape modeling in the \textit{top-down} fashion can provide strong shape guidance for segmentation. This kind of shape guidance is crucial for bridging the gaps on boundary in ultrasound images, but previous methods tend to consider the shape modeling in a static manner, and handle the landmark descriptor design and shape prior extraction separately. In this paper, we propose to consider prostate ultrasound image segmentation as a dynamic and sequential procedure, and complete the descriptor learning and shape inference simultaneously. \par

Our framework has three key modules. \textit{Firstly}, we propose to serialize static prostate ultrasound images into dynamic sequences and then infer prostate shapes by exploring shape priors sequentially. This interpretation is natural and can be formulated with the biologically plausible Recurrent Neural Networks (RNNs) \cite{hochreiter1997long}. We denote the RNN for this simulation as Boundary Completion RNN (\textit{BCRNN}). Without hand-crafted feature design which is required in traditional shape modeling, our BCRNN directly takes raw intensities as input at each timestep. All descriptors for shape inference can be automatically learned by BCRNN. Inherently, BCRNN is able to access previous timesteps through hidden states and exploits them as shape knowledge to reason current missing parts. Therefore, BCRNN can be utilized to infer the incompleteness along prostate boundary in ultrasound image, and thus boost the segmentation accuracy. \textit{Secondly}, we observed that serializing static ultrasound image with different starting points causes bias in shape prediction. In this regard, we adopt a multi-view fusion strategy to merge shape predictions obtained from different perspectives into a comprehensive prediction. \textit{Thirdly}, to combine hierarchical cues of boundary predictions, we further implant our BCRNN core into a multiscale Auto-Context scheme to gain incremental refinement on details of prostate shape prediction. In this scheme, BCRNN with fine scale can dramatically benefit from the shape predictions provided by BCRNN with coarse scale. In addition, different from the traditional classifiers used in Auto-Context scheme \cite{tu2010auto}, our BCRNN can flexibly leverage contextual information in sequence without using empirically designed context structure. \par

\section{Methodology}
\begin{figure}
	\centering
	\includegraphics[width=0.9\linewidth]{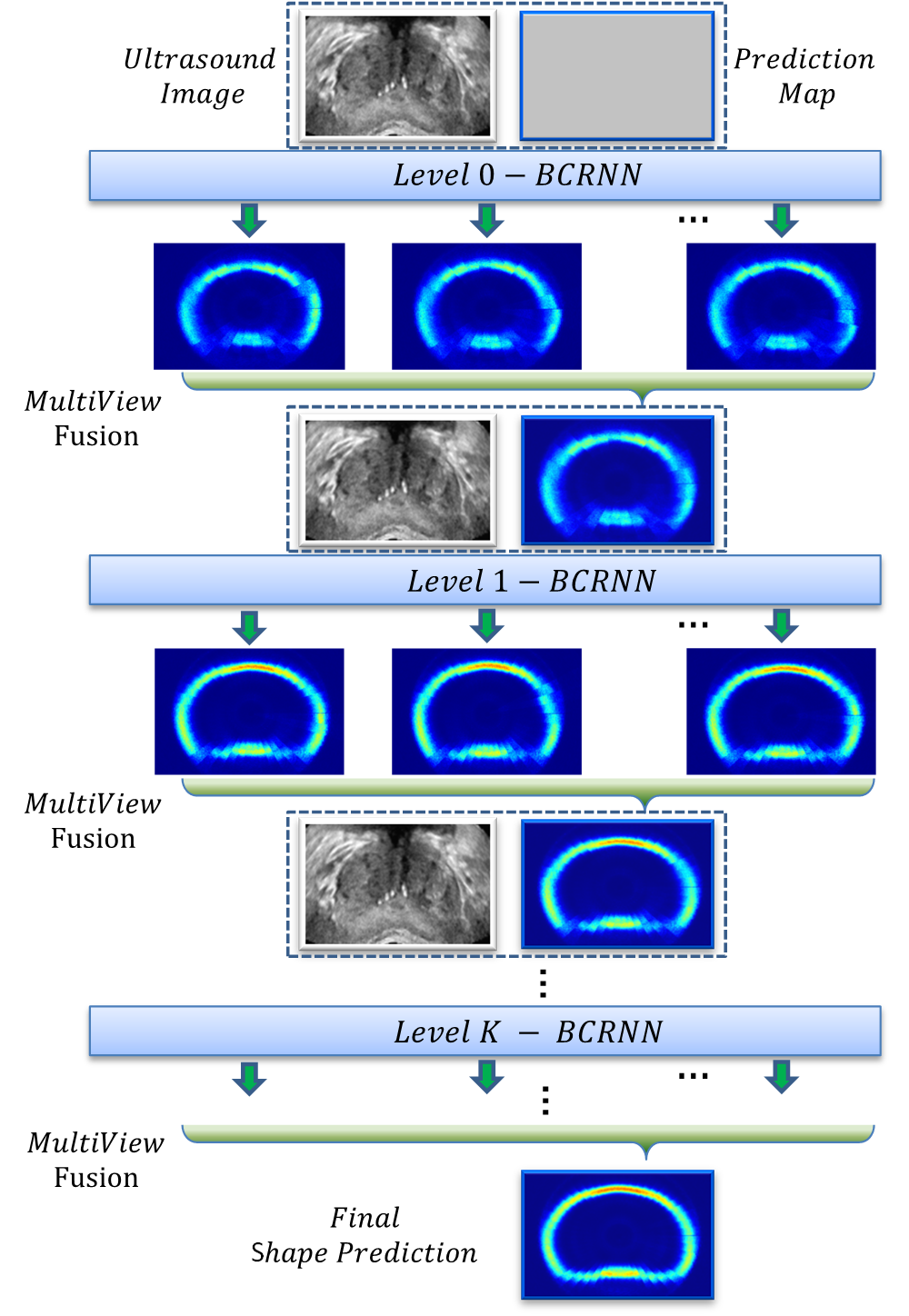}
	\caption{Illustration of our proposed framework. The shape predictions present to be gradually refined as the original prostate ultrasound image flows through all context levels.}
	\label{fig:Framework}
\end{figure}

Our proposed framework is illustrated in Fig.~\ref{fig:Framework}. The core of our framework is the BCRNN in each cascaded level. The cropped prostate ultrasound image is the input of our framework. At each level, BCRNN serializes the static ultrasound image from several different perspectives and then conducts shape prediction on these serializations independently. Those shape predictions are then merged as a comprehensive inference by a multi-view fusion strategy. This comprehensive inference result is subsequently concatenated with the original static ultrasound image, and fed into the next level where more shape details are emphasized. The shape prediction map is initially set as an uniform distribution. The whole procedure iterates from levels with coarse scale to levels with fine scale until convergence occurs on boundary prediction map. \par

\subsection{Shape Inference with BCRNN}
The boundary incompleteness in prostate ultrasound image remains to be the most difficult part for automatic prostate segmentation. The work in this paper is mainly motivated by previous studies about boundary completion. \par

\begin{figure}
	\centering
	\includegraphics[width=1.0\linewidth]{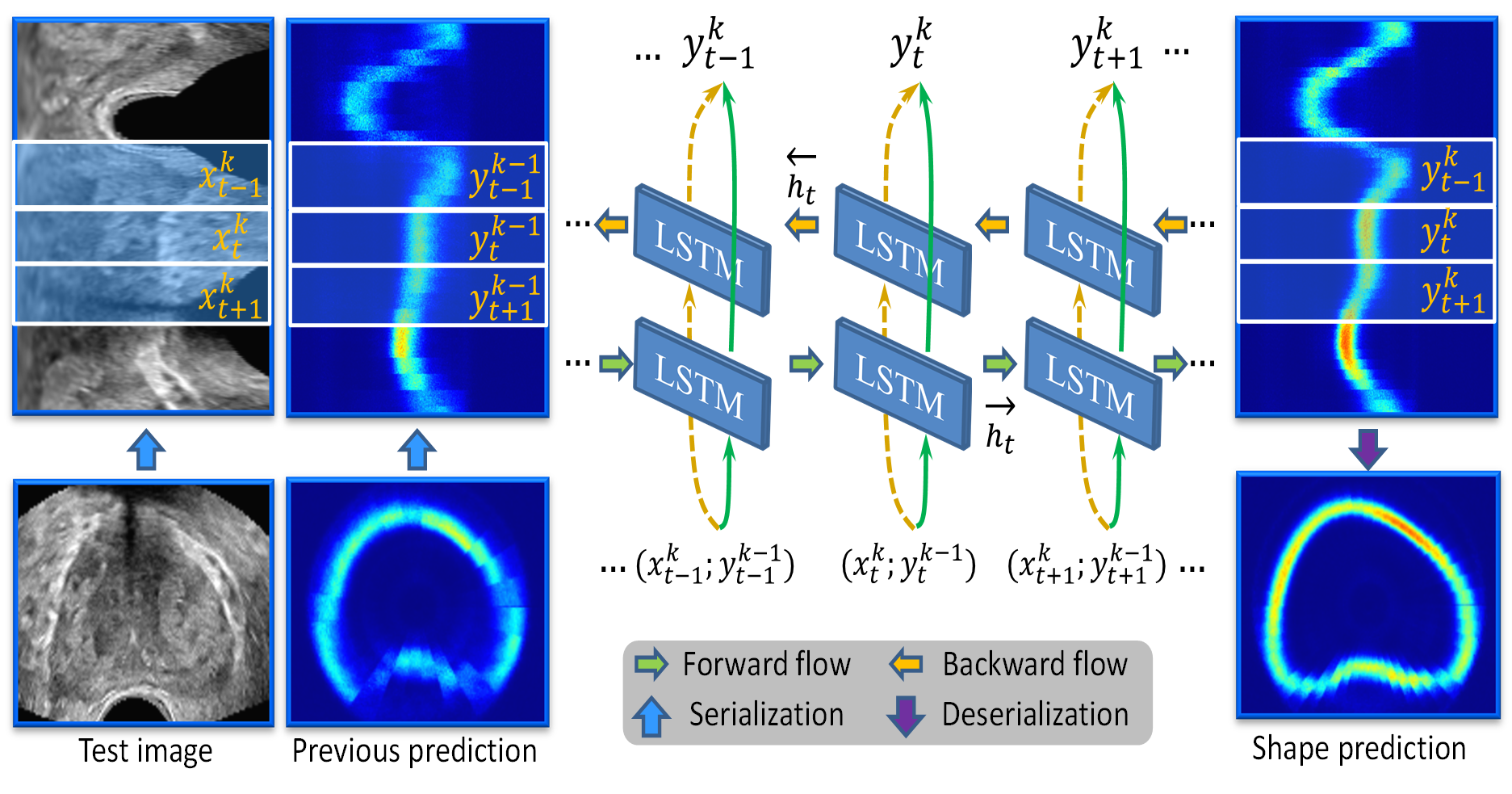}
	\caption{Structure of a Bidirectional LSTM based BCRNN.}
	\label{fig:BCRNN}
\end{figure}

Different from the completion methods which utilize geometric analysis of curvature \cite{kimia2003euler,rueda2015feature} and visual cortex simulation \cite{ben2012tangent}, we propose to formulate the completion problem as a memory guided inference procedure, because we observed that boundary delineation is naturally more like a sequential procedure. Human beings actually leverage indicative inducers from various ranges along object boundary fragments to infer current boundary location, especially when bridging blind spots. Also, this inference process is inherently guided by the shape prior which is memorized by brain and dynamically rectified by current boundary prediction. We find that this procedure has high conformity with the biologically plausible Recurrent Neural Network, which is featured by its memory-based power in learning from sequences. Thus, to the best of our knowledge, for the first time, the Recurrent Neural Network, especially that equipped with the Long Short-Term Memory \cite{hochreiter1997long}, is explored for automatic shape inference (Fig.~\ref{fig:BCRNN}). We denote the RNN for this inference as Boundary Completion RNN (\textit{BCRNN}).

\textbf{Serialization and Deserialization.} Before we can utilize BCRNN for shape inference, we need to transform the static ultrasound image into an interpretable sequence. Among all possible strategies, we choose the strategy which transforms the ultrasound image from Cartesian into polar coordinate system around image center to generate a serialization (shown as Fig.~\ref{fig:BCRNN}). This strategy is straightforward and mainly motivated by the circle-wise manner of manual delineation. The serialization result is still in image form. Deserialization is the inverse process. The serialization image is then evenly partitioned into $T$ consecutive bands which finally form the sequence $\bm{x}=(x_{1},...,x_{T})$ . The flattened version of $\bm{x}$ is then sequentially input into a BCRNN as timestep increases. Specifically, we transformed all ultrasound images into serialization images with the size 400$\times$400 pixels. In BCRNN training, we also serialized the segmentation label images into sequence form. Note that our segmentation algorithm is not that sensitive to object position shift, which means slight offset caused by object localization is allowed during the serialization process. \par

\textbf{Recurrent Neural Networks.} Given an input sequence $\bm{x}$, a basic RNN computes the hidden state vector $\bm{h}=(h_{1},...,h_{T})$ and output vector $\bm{y}=(y_{1},...,y_{T})$ by iterating the following equations from timestep $t = 1$ to $T$:
\begin{align}
\label{eq:rnn_h_t}
h_{t} & = \mathcal{H}(W_{xh}x_{t} + W_{hh}h_{t-1} + b_{h}) \; \\
\label{eq:rnn_y_t}
y_{t} & = W_{hy}h_{t} + b_{y}
\end{align}
where the $W$ terms denote network weight matrices, the $b$ terms denote bias vectors and $\mathcal{H}$ is the hidden layer function. Since hidden state vector $h_{t}$ summarizes the information from previous $h_{t-1}$ and current input $x_{t}$, it can be exploited to infer current prediction $y_{t}$. For the prostate segmentation problem, the hidden state vector $h_{t}$ can be considered as shape knowledge about prostate which is accumulated from the segmentation in previous timesteps. The RNN can use $h_{t}$ to infer the boundary location for current timestep $t$ by taking an extra input $x_{t}$. \par

One problem with basic RNNs is that they are incapable of accessing long-range context, because the information stored in hidden layer tends to decay over time and therefore gradually loses impact on future inference. The invention of Long Short-Term Memory (\textit{LSTM}) module addresses this issue and enhances $\mathcal{H}$ by adding tunable gating units which allow the network to control the flow of information in and out of the network memory. \par

Boundary information from multiple directions are crucial in estimating missing parts. However, a single LSTM stream can only make use of contextual information from one direction. Bidirectional LSTM (\textit{BiLSTM}) addresses this problem by leveraging historical and future information simultaneously. BiLSTM processes sequences from opposite directions with two separate hidden layers, which are then fed forward to the same output layer \cite{graves2013hybrid}. Eq.~\ref{eq:forward}-\ref{eq:combine} present the core computation of a BiLSTM at timestep $t$. As illustrated in Fig.~\ref{fig:BCRNN}, BiLSTM computes forward hidden state $\overrightarrow{h}_{t}$ and backward hidden state $\overleftarrow{h}_{t}$ by iterating the forward layer from t = 1 to $t$, the backward layer from t = $T$ to $t$, respectively.
\begin{align}
\label{eq:forward}
\overrightarrow{h}_{t} & = \mathcal{H}(W_{x\overrightarrow{h}}x_{t} + W_{\overrightarrow{h}\overrightarrow{h}}\overrightarrow{h}_{t-1} + b_{\overrightarrow{h}}) \; \\
\label{eq:backward}
\overleftarrow{h}_{t} & = \mathcal{H}(W_{x\overleftarrow{h}}x_{t} + W_{\overleftarrow{h}\overleftarrow{h}}\overleftarrow{h}_{t-1} + b_{\overleftarrow{h}}) \; \\
\label{eq:combine}
y_{t} & = W_{\overrightarrow{h}y}\overrightarrow{h}_{t} + W_{\overleftarrow{h}y}\overleftarrow{h}_{t} + b_{y}
\end{align}
As shown in Eq.~\ref{eq:combine}, interaction happens between forward hidden state $\overrightarrow{h}_{t}$ and backward hidden state $\overleftarrow{h}_{t}$ in BiLSTM. By combining serialization and deserialization with a BiLSTM, we can build a BCRNN to properly blend boundary clues from multiple directions. This memory based BCRNN presents great advantages in estimating incomplete boundaries of prostate in ultrasound images.

\subsection{Multiple Viewpoint Fusion}
\begin{figure}
	\centering
	\includegraphics[width=1.0\linewidth]{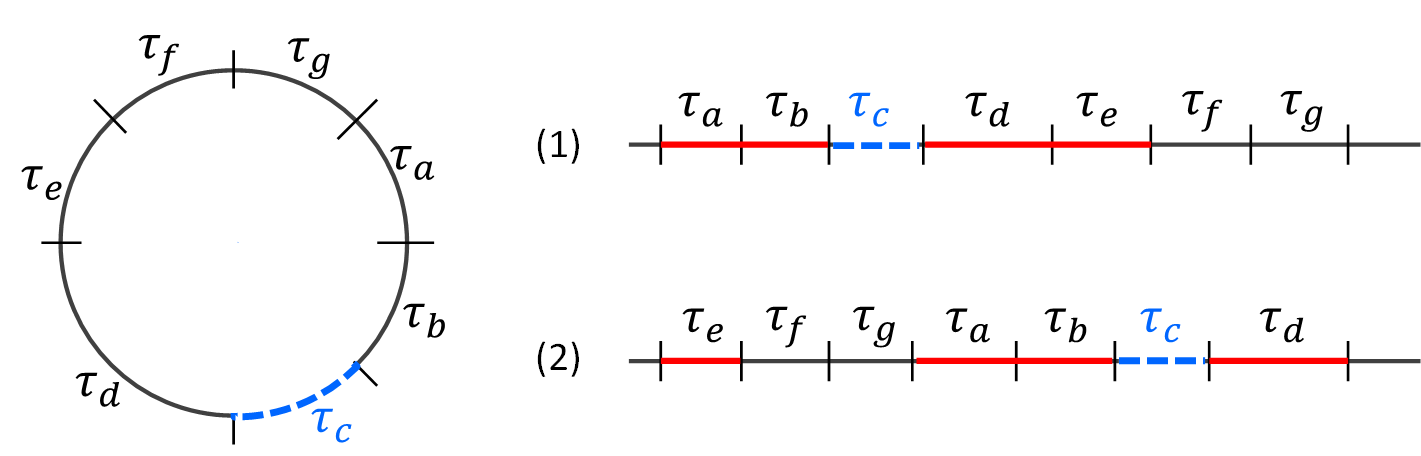}
	\caption{Serialization manners with different starting points. Serialization (1) from $\tau_{a}$ preserves the contextual dependency of $\tau_{c}$, while serialization (2) from $\tau_{e}$ destroys the dependency by pushing $\tau_{e}$ faraway from $\tau_{d}$ and $\tau_{c}$.}
	\label{fig:starting_points}
\end{figure}

In practice, we find that serializing static ultrasound image from different starting points will cause slightly different shape predictions. We interpret this phenomenon as that serializing from different starting points may change the relative distances between context-dependent sequence elements, and therefore brings about slight difference in predictions. Shown as Fig. \ref{fig:starting_points}, suppose $\tau_{c}$ is the missing boundary fragment and we mainly need clues from both $\tau_{a-b}$ and $\tau_{d-e}$ fragments to recover it. The first serialization manner preserves the relative spatial relationship between these three fragments, while the second manner destroys the continuity between $\tau_{e}$ and $\tau_{d}$, and makes $\tau_{e}$ become far away from $\tau_{c}$. In the second case, information about $\tau_{e}$ needs to be kept much longer by BiLSTM before it achieves $\tau_{c}$, which is more challenging under limited memory unit resources. \par

To solve this problem efficiently, we choose to serialize the original static ultrasound image from three different viewpoints, and then democratically merge the three complementary boundary predictions which are generated by the same BCRNN into a comprehensive shape prediction. Because this fusion procedure is similar as human does in getting the impression of an object by observing from multiple viewpoints, we denote it as multiple viewpoint fusion. \par

\subsection{Multiscale Auto-Context for Refinement}
To enhance spatial consistency and boundary details within the shape prediction map generated by BCRNN, we propose to further implant the BCRNN into a multiscale Auto-Context scheme \cite{tu2010auto}, which can gain successive refinement on the preliminary prediction result by exploring prediction information from neighbors. Specifically, we directly concatenate the prediction map generated by BCRNN in level $k-1$ with the original ultrasound image, and take them as the input for BCRNN in level $k$. Also, the BCRNN model in level $k$ is only trained after the training of BCRNN in level $k-1$ has finished. Traditional classifiers implanted into Auto-Context scheme often rely on some empirically designed structures to collect contextual information \cite{tu2010auto,gao2014learning}, while our BCRNN has the inbuilt ability to flexibly leverage context information from near or far ranges. This ability benefits from the memory which is retained dynamically by BiLSTM. \par

In practice, it's difficult to decide the optimal scale of $x_{t}$ when splitting the serialization image into a sequence $\bm{x}$. Large scale $x_{t}$ suppresses the boundary details as one timestep, while small scale $x_{t}$ becomes less informative and makes the sequence tediously long. Motivated by the fact that detailed boundary delineations are often conducted after a coarse sketch is obtained, we configure our BCRNN embedded Auto-Context scheme with a multi-scale mechanism. In this mechanism, BCRNNs in early levels with large scales can only produce coarse shape prediction maps, but those informative maps can provide strong guidance for BCRNNs in levels with fine scales. Eq.~\ref{eq:autocontext} formulates the iterative process of our multiscale Auto-Context scheme, where $\mathcal{F}$ is the BCRNN model function; $\bm{y}^{k}$ is the shape prediction map from level $k$; $s^{k}$ is the scale used by BCRNN in level $k$ to generate sequence $\bm{x}^{k}$ (i.e., the height of the band $x_{t}^{k}$ in Fig. \ref{fig:BCRNN}). $y_{t}^{k}$ has the same size with $x_{t}^{k}$. Three context levels are adopted in this paper, with $s^{0}=16$, $s^{1}=10$ and $s^{2}=8$ respectively.
\begin{align}
\label{eq:autocontext}
\bm{y}^{k} & = \mathcal{F}((\bm{x}^{k}; y^{k-1}), s^{k})
\end{align}

Although the cascaded multiscale BCRNNs are robust in recovering missing boundary, currently there is no theoretical guarantee that they can recover all missing boundaries with an absolutely close form. So, after the last BCRNN level, we propose to apply an auxiliary ASM model \cite{van2002active} on the  shape prediction map to generate the final segmentation. This auxiliary ASM model is built on 300 annotated prostate shape maps. These maps are obtained by running our BCRNNs on 300 prostate ultrasound images in training dataset, and each map is evenly annotated with 12 main and 60 secondary landmarks. Although only intensity information of the map is used to describe each landmark, this ASM model has very little chance to be corrupted by local boundary uncertainty, because it becomes much easier for the model to fit prostate shape in the prediction map than that in original ultrasound image. Also, since most ambiguous and long-span occluded boundaries are recovered by our cascaded BCRNNs, only a few small gaps are left for ASM model to bridge (Fig. \ref{fig:quality_show}).

\begin{table*}[!htb] \caption {Quantitative evaluation of different methods}
	\label{table:results_comparison}
	\centering
	\begin{tabular}{c|c|c|c|c|c|c}
		\toprule[2pt]
		\bf{Method} 			& Dice 	 	& Adb 		& Conform	& Jaccard 	& Precision	 	& Recall\\
		\hline
		\textit{T-CNN} 			&0.9206		&12.7312	&0.8251		&0.8541		&0.8966		&\textbf{0.9495}\\
		\textit{T-FCN} 		&0.9188		&12.6720	&0.8207		&0.8513		&0.9334		&0.9080\\
		\hline
		BCRNN-\textit{Level0}	&0.9091		&14.0688	&0.7975		&0.8348		&0.9286		&0.8921 \\
		BCRNN-\textit{Level1}	&\textbf{0.9239}		&11.5903	&\textbf{0.8322}		&\textbf{0.8602}		&0.9446		&0.9051 \\
		BCRNN-\textit{Level2}	&0.9233		&\textbf{11.4456}	&0.8306		&0.8595		&\textbf{0.9519}		&0.8976 \\
		
		\toprule[2pt]
	\end{tabular}
\end{table*}

\section{Experimental Results}
\subsection{Materials and Preprocessing}
We collected 17 trans-rectal ultrasound (TRUS) volumes which were acquired from 17 patients. These 3D TRUS volumes were obtained by a Mindray DC-8 ultrasound system with an integrated 3D TRUS probe. The size of 3D TRUS volume is 214$\times$125$\times$44 with a voxel size of 0.5$\times$0.5$\times$0.5 mm. We totally extracted all 530 slices which contain prostate from those volumes, and augmented 400 slices of 10 patients to 2400 images as training set, the rest 130 slices from 7 patients were taken as testing set. An experienced radiologist provided segmentation labels for all images. Because the basic assumption of our method is that, the object to be segmented is located around the center of field of view, so the input of our method is the cropped image region of prostate, and the automatic prostate localization in ultrasound image is out of the scope of this paper. \par

\subsection{Implementation Details}
Our proposed framework was implemented with the popular library \textit{Theano} \cite{al2016theano}. We trained each BCRNN with a many-to-many manner, so that a direct mapping was built between the input raw intensity sequence and the boundary label sequence. The forward and backward LSTM streams in BCRNN contain 500 hidden memory units, respectively. There was no pre-training for our network and all weights were randomly initialized from a Gaussian distribution. We trained each BCRNN by minimizing an Euclidean distance based objective function and iteratively updated the network parameters with RMSProp optimizer \cite{tieleman2012rmsprop} using the backpropagation through time algorithm (BPTT). The learning rate was set as 0.001 for all context levels and about 2 hours were needed to train each level. All computations were conducted on a computer equipped with dual Intel Xeon(R) processors E5-2650 2.6 GHz and a GPU of NVIDIA GeForce GTX TITAN X. \par

\subsection{Learning Process Analysis}
\begin{figure}
	\centering
	\includegraphics[width=1.0\linewidth]{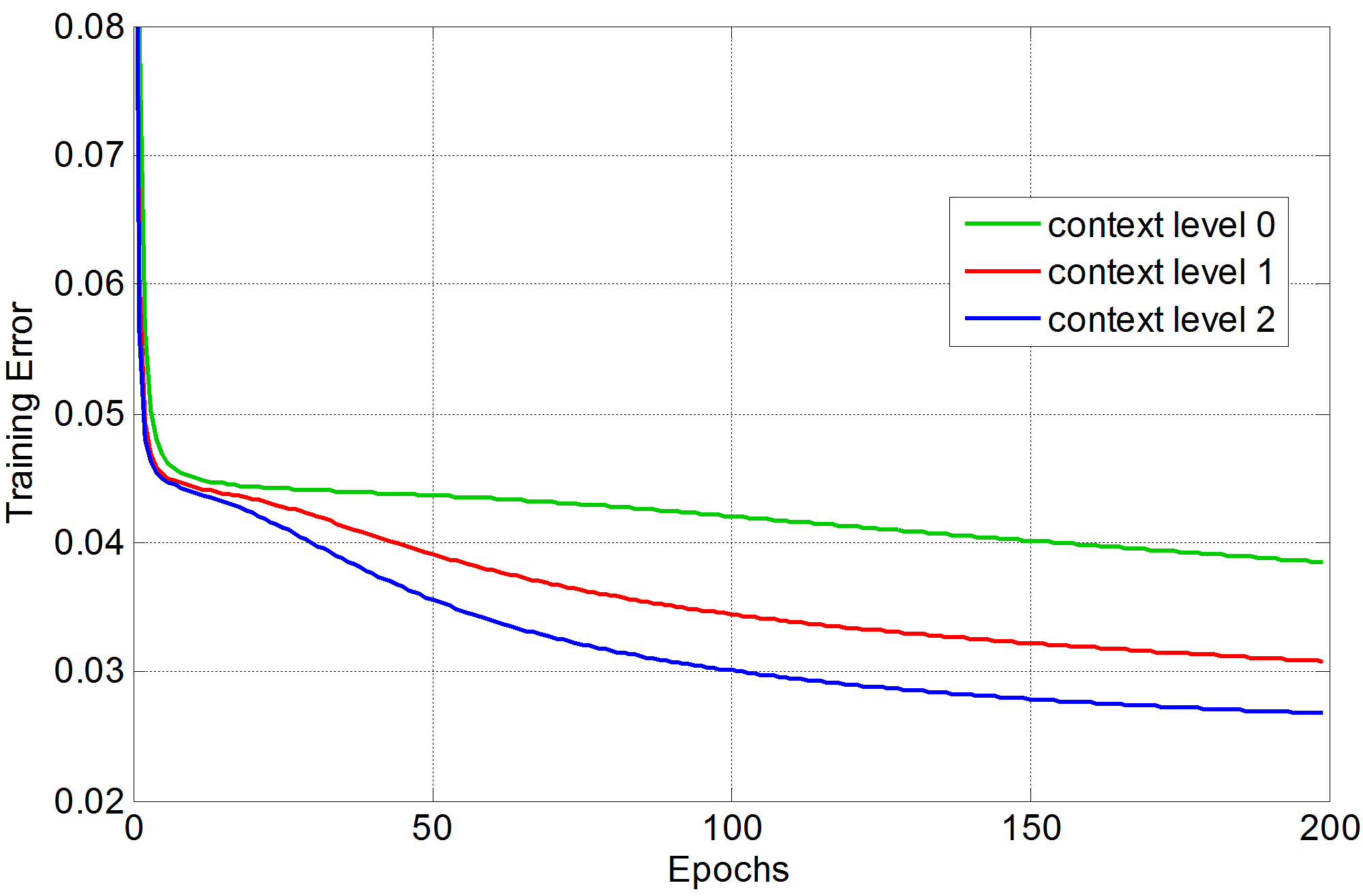}
	\caption{Comparison of the learning curves from successive BCRNNs.}
	\label{fig:errfor3levels}
\end{figure}

It is observed from Fig.~\ref{fig:errfor3levels} that the training error of BCRNN at context level 0 with coarse scale decreases sluggishly after 50 epochs, while the training errors of BCRNNs at context level 1 and 2 with fine scales decrease steeply from very early epochs. This demonstrates that the training of BCRNNs at lower levels facilitates the training of BCRNNs at higher levels. This is because the prediction maps generated by lower levels provide strong guidance for following levels and consequently accelerate the optimization of network parameters. \par

\subsection{Qualitative Evaluation}
\begin{figure*}
	\centering
	\includegraphics[width=1.0\linewidth]{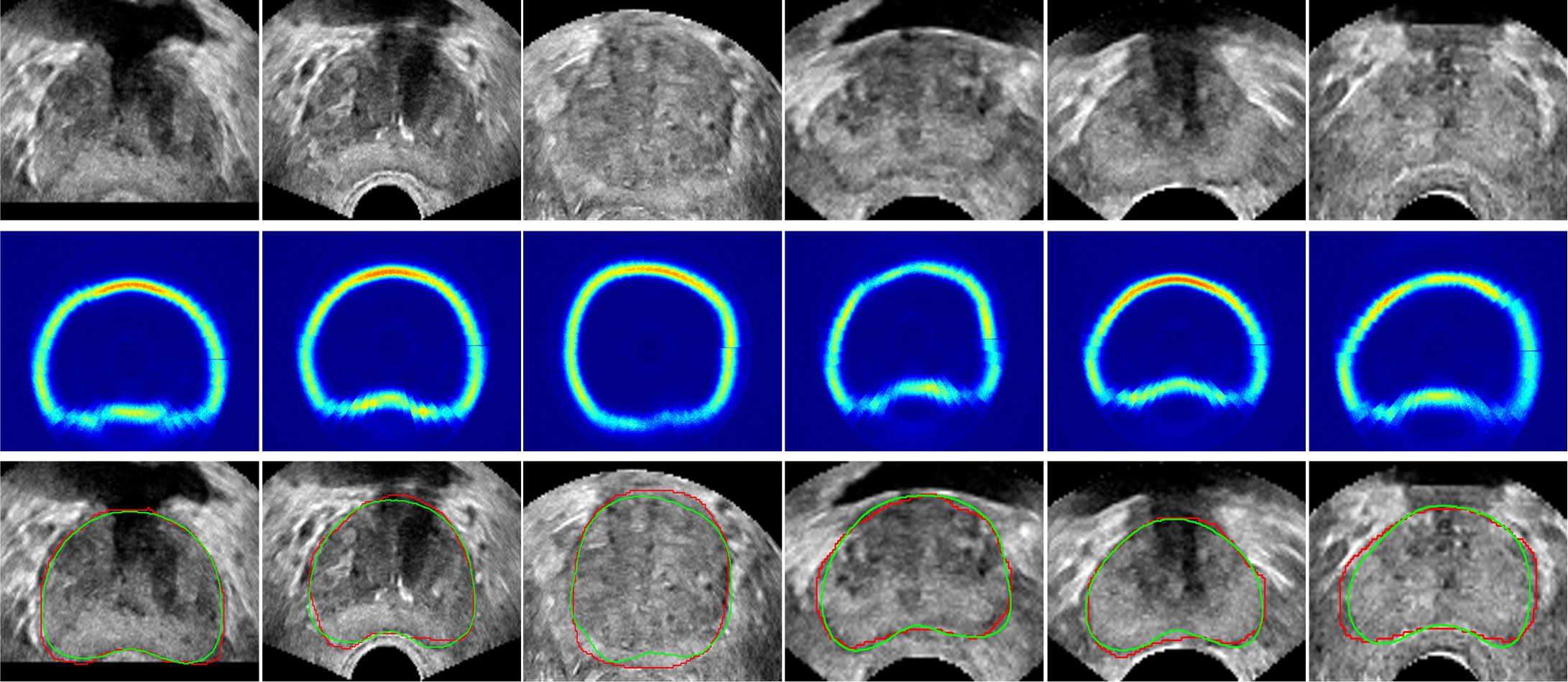}
	\caption{Our results on prostate segmentation in trans-rectal ultrasound images. From top to bottom: prostate ultrasound image, shape prediction and segmentation result. Green and red curves denote automatic segmentation and ground truth, respectively.}
	\label{fig:quality_show}
	\vspace{-0.5cm}
\end{figure*}

In Fig.~\ref{fig:quality_show}, we illustrate the prostate segmentation results of our method along with the shape prediction maps produced by BCRNN at level 2. By simultaneously learning boundary descriptors and exploring sequential information for inference, our method can not only successfully infer ambiguous and deficient boundaries in low contrast prostate ultrasound images, but also conquer the large inter-variance of prostate shape and size. Importantly, our method is robust in distinguishing the inhomogeneous prostate tissues, and recognizing them as a whole part. The auxiliary ASM model proves to work well under the strong guidance provided by shape prediction map. \par

\subsection{Quantitative Evaluation}
Metrics evaluating area and shape similarities are both adopted, such as Dice Similarity Coefficient (Dice), Average Distance of Boundaries (Adb [pixel unit]), Conformity (Conform), Jaccard Index (Jaccard), Precision and Recall. We extensively compared our method with several advanced methods, including Convolutional Neural Network (CNN) \cite{krizhevsky2012imagenet} and Fully Convolutional Network (FCN) \cite{long2015fully}. It should be noted that, our BCRNNs were trained from scratch, while the compared CNN was pre-trained on other ultrasound images and FCN was transferred from VGG16 model \cite{simonyan2014very} which was trained on Imagenet dataset \cite{deng2009imagenet}. The pre-trained CNN and FCN are denoted as T-CNN and T-FCN. \par

Table~\ref{table:results_comparison} illustrates the detailed comparison between different methods, as well as the comparison between different BCRNN levels. We can observe that, benefiting from the prediction result from BCRNN-\textit{Level0}, BCRNN-\textit{Level1} gains considerable improvement on all evaluation metrics. Although the BCRNN-\textit{Level0} at coarse level performs worse than \textit{T-CNN} and \textit{T-FCN}, BCRNN-\textit{Level1} and BCRNN-\textit{Level2} present very competitive and even better results than \textit{T-CNN} and \textit{T-FCN} in key metrics. Generally, the refinement contributed by Auto-Context scheme diminishes exponentially as context level increases, and stacking too many context levels may lead to overfitting and performance drop. So in our case, we only adopt three BCRNN levels, because we can observe that the improvement from BCRNN-\textit{Level1} to BCRNN-\textit{Level2} already presents to be marginal. \par

\section{Conclusion}
In this paper, we propose a biologically plausible method to combat the boundary incompleteness challenge for automatic prostate segmentation in ultrasound image. We creatively formulate the boundary completion as a sequential problem. Originating from RNNs, our intuitive method dynamically explores sequential clues about past and future to learn the shape knowledge. Combining with a multiscale Auto-Context scheme further offers us opportunities to enhance shape prediction details. Our method presents intriguing abilities in recovering severe incompleteness, as demonstrated in the challenging prostate ultrasound images. \par

\section{Acknowledgments}
The work described in this paper was supported by the grant from the National Basic Program of China, 973 Program (Project No. 2015CB351706), the grant from the Research Grants Council of the Hong Kong Special Administrative Region (Project no. CUHK 14202514) and the grant from the Hong Kong-Shenzhen Innovation Circle Funding Program (No. GHP/002/13SZ and SGLH20131010151755080). \par

\bibliographystyle{aaai}
\bibliography{aaai2017_ps}

\begin{thebibliography}{}

\bibitem[\protect\citeauthoryear{Al-Rfou \bgroup et al\mbox.\egroup
  }{2016}]{al2016theano}
Al-Rfou, R.; Alain, G.; Almahairi, A.; Angermueller, C.; et~al.
\newblock 2016.
\newblock Theano: A python framework for fast computation of mathematical
  expressions.
\newblock {\em arXiv eprints abs/1605.02688 (May 2016). url: http://arxiv.
  org/abs/1605.02688}.

\bibitem[\protect\citeauthoryear{Ben-Yosef and
  Ben-Shahar}{2012}]{ben2012tangent}
Ben-Yosef, G., and Ben-Shahar, O.
\newblock 2012.
\newblock A tangent bundle theory for visual curve completion.
\newblock {\em IEEE transactions on pattern analysis and machine intelligence}
  34(7):1263--1280.

\bibitem[\protect\citeauthoryear{Chen \bgroup et al\mbox.\egroup
  }{2016}]{chen2016iterative}
Chen, H.; Zheng, Y.; Park, J.-H.; Heng, P.-A.; and Zhou, S.~K.
\newblock 2016.
\newblock Iterative multi-domain regularized deep learning for anatomical
  structure detection and segmentation from ultrasound images.
\newblock In {\em International Conference on Medical Image Computing and
  Computer-Assisted Intervention},  487--495.
\newblock Springer.

\bibitem[\protect\citeauthoryear{Cootes \bgroup et al\mbox.\egroup
  }{1995}]{cootes1995active}
Cootes, T.~F.; Taylor, C.~J.; Cooper, D.~H.; and Graham, J.
\newblock 1995.
\newblock Active shape models-their training and application.
\newblock {\em Computer vision and image understanding} 61(1):38--59.

\bibitem[\protect\citeauthoryear{Deng \bgroup et al\mbox.\egroup
  }{2009}]{deng2009imagenet}
Deng, J.; Dong, W.; Socher, R.; Li, L.-J.; Li, K.; and Fei-Fei, L.
\newblock 2009.
\newblock Imagenet: A large-scale hierarchical image database.
\newblock In {\em Computer Vision and Pattern Recognition, 2009. CVPR 2009.
  IEEE Conference on},  248--255.
\newblock IEEE.

\bibitem[\protect\citeauthoryear{Gao \bgroup et al\mbox.\egroup
  }{2014}]{gao2014learning}
Gao, Y.; Wang, L.; Shao, Y.; and Shen, D.
\newblock 2014.
\newblock Learning distance transform for boundary detection and deformable
  segmentation in ct prostate images.
\newblock In {\em International Workshop on Machine Learning in Medical
  Imaging},  93--100.
\newblock Springer.

\bibitem[\protect\citeauthoryear{Ghose \bgroup et al\mbox.\egroup
  }{2013}]{ghose2013supervised}
Ghose, S.; Oliver, A.; Mitra, J.; Mart{\'\i}, R.; Llad{\'o}, X.; Freixenet, J.;
  Sidib{\'e}, D.; Vilanova, J.~C.; Comet, J.; and Meriaudeau, F.
\newblock 2013.
\newblock A supervised learning framework of statistical shape and probability
  priors for automatic prostate segmentation in ultrasound images.
\newblock {\em Medical image analysis} 17(6):587--600.

\bibitem[\protect\citeauthoryear{Graves, Jaitly, and
  Mohamed}{2013}]{graves2013hybrid}
Graves, A.; Jaitly, N.; and Mohamed, A.-r.
\newblock 2013.
\newblock Hybrid speech recognition with deep bidirectional lstm.
\newblock In {\em Automatic Speech Recognition and Understanding (ASRU), 2013
  IEEE Workshop on},  273--278.
\newblock IEEE.

\bibitem[\protect\citeauthoryear{Guo, Gao, and Shen}{2016}]{guo2016deformable}
Guo, Y.; Gao, Y.; and Shen, D.
\newblock 2016.
\newblock Deformable mr prostate segmentation via deep feature learning and
  sparse patch matching.
\newblock {\em IEEE transactions on medical imaging} 35(4):1077--1089.

\bibitem[\protect\citeauthoryear{Hochreiter and
  Schmidhuber}{1997}]{hochreiter1997long}
Hochreiter, S., and Schmidhuber, J.
\newblock 1997.
\newblock Long short-term memory.
\newblock {\em Neural computation} 9(8):1735--1780.

\bibitem[\protect\citeauthoryear{Hodge \bgroup et al\mbox.\egroup
  }{1989}]{hodge1989random}
Hodge, K.; McNeal, J.; Terris, M.~K.; and Stamey, T.
\newblock 1989.
\newblock Random systematic versus directed ultrasound guided transrectal core
  biopsies of the prostate.
\newblock {\em The Journal of urology} 142(1):71--4.

\bibitem[\protect\citeauthoryear{Kimia, Frankel, and
  Popescu}{2003}]{kimia2003euler}
Kimia, B.~B.; Frankel, I.; and Popescu, A.-M.
\newblock 2003.
\newblock Euler spiral for shape completion.
\newblock {\em International journal of computer vision} 54(1-3):159--182.

\bibitem[\protect\citeauthoryear{Krizhevsky, Sutskever, and
  Hinton}{2012}]{krizhevsky2012imagenet}
Krizhevsky, A.; Sutskever, I.; and Hinton, G.~E.
\newblock 2012.
\newblock Imagenet classification with deep convolutional neural networks.
\newblock In {\em Advances in neural information processing systems},
  1097--1105.

\bibitem[\protect\citeauthoryear{Li and Malik}{2016}]{li2016amodal}
Li, K., and Malik, J.
\newblock 2016.
\newblock Amodal instance segmentation.
\newblock {\em arXiv preprint arXiv:1604.08202}.

\bibitem[\protect\citeauthoryear{Long, Shelhamer, and
  Darrell}{2015}]{long2015fully}
Long, J.; Shelhamer, E.; and Darrell, T.
\newblock 2015.
\newblock Fully convolutional networks for semantic segmentation.
\newblock In {\em Proceedings of the IEEE Conference on Computer Vision and
  Pattern Recognition},  3431--3440.

\bibitem[\protect\citeauthoryear{Mahapatra and
  Buhmann}{2015}]{mahapatra2015visual}
Mahapatra, D., and Buhmann, J.~M.
\newblock 2015.
\newblock Visual saliency based active learning for prostate mri segmentation.
\newblock In {\em International Workshop on Machine Learning in Medical
  Imaging},  9--16.
\newblock Springer.

\bibitem[\protect\citeauthoryear{Noble and
  Boukerroui}{2006}]{noble2006ultrasound}
Noble, J.~A., and Boukerroui, D.
\newblock 2006.
\newblock Ultrasound image segmentation: a survey.
\newblock {\em IEEE Transactions on medical imaging} 25(8):987--1010.

\bibitem[\protect\citeauthoryear{Rogers and Graham}{2002}]{rogers2002robust}
Rogers, M., and Graham, J.
\newblock 2002.
\newblock Robust active shape model search.
\newblock In {\em European conference on computer vision},  517--530.
\newblock Springer.

\bibitem[\protect\citeauthoryear{Rueda \bgroup et al\mbox.\egroup
  }{2015}]{rueda2015feature}
Rueda, S.; Knight, C.~L.; Papageorghiou, A.~T.; and Noble, J.~A.
\newblock 2015.
\newblock Feature-based fuzzy connectedness segmentation of ultrasound images
  with an object completion step.
\newblock {\em Medical image analysis} 26(1):30--46.

\bibitem[\protect\citeauthoryear{Santiago, Nascimento, and
  Marques}{2015}]{santiago20152d}
Santiago, C.; Nascimento, J.~C.; and Marques, J.~S.
\newblock 2015.
\newblock 2d segmentation using a robust active shape model with the em
  algorithm.
\newblock {\em IEEE Transactions on Image Processing} 24(8):2592--2601.

\bibitem[\protect\citeauthoryear{Shen, Zhan, and
  Davatzikos}{2003}]{shen2003segmentation}
Shen, D.; Zhan, Y.; and Davatzikos, C.
\newblock 2003.
\newblock Segmentation of prostate boundaries from ultrasound images using
  statistical shape model.
\newblock {\em IEEE transactions on medical imaging} 22(4):539--551.

\bibitem[\protect\citeauthoryear{Simonyan and
  Zisserman}{2014}]{simonyan2014very}
Simonyan, K., and Zisserman, A.
\newblock 2014.
\newblock Very deep convolutional networks for large-scale image recognition.
\newblock {\em arXiv preprint arXiv:1409.1556}.

\bibitem[\protect\citeauthoryear{Terris and
  Stamey}{1991}]{terris1991determination}
Terris, M.~K., and Stamey, T.
\newblock 1991.
\newblock Determination of prostate volume by transrectal ultrasound.
\newblock {\em The Journal of urology} 145(5):984--987.

\bibitem[\protect\citeauthoryear{Tieleman and
  Hinton}{2012}]{tieleman2012rmsprop}
Tieleman, T., and Hinton, G.
\newblock 2012.
\newblock Lecture 6.5-rmsprop: Divide the gradient by a running average of its
  recent magnitude.
\newblock {\em COURSERA: Neural Networks for Machine Learning} 4(2).

\bibitem[\protect\citeauthoryear{Tu and Bai}{2010}]{tu2010auto}
Tu, Z., and Bai, X.
\newblock 2010.
\newblock Auto-context and its application to high-level vision tasks and 3d
  brain image segmentation.
\newblock {\em IEEE Transactions on Pattern Analysis and Machine Intelligence}
  32(10):1744--1757.

\bibitem[\protect\citeauthoryear{Van~Ginneken \bgroup et al\mbox.\egroup
  }{2002}]{van2002active}
Van~Ginneken, B.; Frangi, A.~F.; Staal, J.~J.; ter Haar~Romeny, B.~M.; and
  Viergever, M.~A.
\newblock 2002.
\newblock Active shape model segmentation with optimal features.
\newblock {\em IEEE transactions on medical imaging} 21(8):924--933.

\bibitem[\protect\citeauthoryear{Wang \bgroup et al\mbox.\egroup
  }{2016}]{wang2016towards}
Wang, Y.; Cheng, J.-Z.; Ni, D.; Lin, M.; Qin, J.; Luo, X.; Xu, M.; Xie, X.; and
  Heng, P.~A.
\newblock 2016.
\newblock Towards personalized statistical deformable model and hybrid point
  matching for robust mr-trus registration.
\newblock {\em IEEE transactions on medical imaging} 35(2):589--604.

\bibitem[\protect\citeauthoryear{Yan \bgroup et al\mbox.\egroup
  }{2010}]{yan2010discrete}
Yan, P.; Xu, S.; Turkbey, B.; and Kruecker, J.
\newblock 2010.
\newblock Discrete deformable model guided by partial active shape model for
  trus image segmentation.
\newblock {\em IEEE Transactions on Biomedical Engineering} 57(5):1158.

\bibitem[\protect\citeauthoryear{Zhan and Shen}{2006}]{zhan2006deformable}
Zhan, Y., and Shen, D.
\newblock 2006.
\newblock Deformable segmentation of 3-d ultrasound prostate images using
  statistical texture matching method.
\newblock {\em IEEE Transactions on Medical Imaging} 25(3):256--272.

\bibitem[\protect\citeauthoryear{Zhou \bgroup et al\mbox.\egroup
  }{2013}]{zhou2013active}
Zhou, X.; Huang, X.; Duncan, J.~S.; and Yu, W.
\newblock 2013.
\newblock Active contours with group similarity.
\newblock In {\em Proceedings of the IEEE Conference on Computer Vision and
  Pattern Recognition},  2969--2976.

\end{thebibliography}

\end{document}